\PassOptionsToPackage{table}{xcolor}
\documentclass[runningheads]{llncs}
\usepackage[T1]{fontenc}

\usepackage{iciap}
\usepackage{iciapabbrv}
\usepackage[accsupp]{axessibility}

\usepackage{hyperref}
\usepackage{orcidlink}   
\usepackage{graphicx}
\usepackage{booktabs}
\usepackage{amsmath}
\usepackage{amssymb}

\usepackage{soul}
\begin{document}

\title{A Real-Time System for Egocentric Hand-Object Interaction Detection in Industrial Domains}

\titlerunning{Real-Time Egocentric HOI Detection in Industrial Settings}

\author{Antonio Finocchiaro\textsuperscript{1,*} \and 
Alessandro Sebastiano Catinello\textsuperscript{1,*} \and 
Michele Mazzamuto\textsuperscript{1}\orcidID{0009-0006-3466-5555} \and 
Rosario Leonardi\textsuperscript{1}\orcidID{0009-0001-8693-3826} \and 
Antonino Furnari\textsuperscript{1}\orcidID{0000-0001-6911-0302} \and  
Giovanni Maria Farinella\textsuperscript{1}\orcidID{0000-0002-6034-0432}}

\authorrunning{Finocchiaro et al.}

\institute{
\textsuperscript{1}Department of Mathematics and Computer Science, University of Catania, Italy \\
\textsuperscript{*}These authors share first authorship \\
\email{1000006272@studium.unict.it, ale.catinello.c@gmail.com, michele.mazzamuto@phd.unict.it, \{rosario.leonardi, antonino.furnari, giovanni.farinella\}@unict.it} \\
\url{https://iplab.dmi.unict.it/fpv/}
}

\maketitle

\begin{abstract}

Hand-object interaction detection remains an open challenge in real-time applications, where intuitive user experiences depend on fast and accurate detection of interactions with surrounding objects. We propose an efficient approach for detecting hand-objects interactions from streaming egocentric vision that operates in real time. Our approach consists of an action recognition module and an object detection module for identifying active objects upon confirmed interaction. Our Mamba model with EfficientNetV2 as backbone for action recognition achieves $38.52\%$ p-AP on the ENIGMA-51 benchmark at $30$fps, while our fine-tuned YOLOWorld reaches $85.13\%$ AP for hand and object.
We implement our models in a cascaded architecture where the action recognition and object detection modules operate sequentially. When the action recognition predicts a contact state, it activates the object detection module, which in turn performs inference on the relevant frame to detect and classify the active object.

\end{abstract}

\keywords{Hand-Object Interaction \and Online Video Understanding \and Wearable Vision.}

\section{Introduction}
Hand-Object Interaction (HOI) detection is a fundamental challenge, particularly when both high accuracy and low latency are required. In industrial scenarios, such as those involving laboratory tools, detecting these interactions becomes crucial, as discussed in \cite{LEONARDI2024103984}. A straightforward method for HOI detection in 2D frames involves detecting hands and objects and applying an Intersection over Union (IoU) threshold to infer contact based on bounding box overlap. While state-of-the-art models exist for HOI detection,  \cite{LEONARDI2024103984, Shan2020}, they are typically computationally intensive and slow. For this reason, we propose a simple yet effective approach that addresses these limitations. Our method leverages the real-time capabilities of detectors like YOLOWorld \cite{cheng2024yolo} to identify active objects. Despite its simplicity, the approach proves to be both effective and lightweight. Moreover, when coupled with an Action Recognition (AR) module, it becomes even more efficient by avoiding per-frame inference.
Finally, we developed an HOI framework which leverages the Meta Quest 3 device. In this implementation, Mamba~\cite{gu2023mamba} was used as the AR module, while YOLOWorld was employed as the object detector to handle active object retrieval, analyzing the overlap between detected hands and objects.


\subsection{Problem Formulation}

As previously defined, the core challenge of this work is the real-time detection of HOIs in egocentric video streams captured in industrial environments.
The problem involves taking streaming video as input and producing localized ``events'' indicating object interactions as output. To be useful in industrial scenarios, the system must perform in real-time.
The performance of the AR module is evaluated using point-level AP (p-AP), as in \cite{shou2018online}. The OD module is evaluated using standard metrics, such as average precision (AP), recall, and harmonic mean (HM). HOI metrics are adopted from \cite{LEONARDI2024103984}.
Both modules take RGB frames as input. The OD module predicts hand coordinates, hand side, active object coordinates, and object class based on the overlap between bounding boxes.
The system behavior depends on the AR module's output, resulting in the following cases. For example, if the AR module predicts a contact but no overlap is detected, the OD module outputs no contact. Conversely, if the AR module predicts a contact and the OD module detects an overlap, the contact state provided by the AR module is used alongside the information retrieved by the OD module.

\section{Related Works}
The practical application of HOI is central to developing effective Mixed Reality (MR) systems for user assistance on wearable devices. Recent work, for example, demonstrates how to support industrial workers by inferring interactions from 2D object detectors and standard MR features, thereby highlighting the need for efficient, real-time models~\cite{mazzamuto2023wearable}.
Egocentric HOI detection focuses on identifying interactions with objects from a first-person perspective, requiring diverse annotation types ranging from basic bounding boxes to multimodal labels including hand detection, contact state estimation, active object identification, segmentation masks, and temporal interaction boundaries. Early work \cite{Shan2020} framed HOI detection as a multitask problem involving hand/object localization and contact-state inference, which was extended to egocentric settings through the MECCANO dataset for industrial applications \cite{ragusa2021meccano} and the EgoHOS benchmark for pixel-level segmentation \cite{zhang2022fine}. The current standard was established with the EPIC-KITCHENS VISOR dataset \cite{darkhalil2022epic}, unifying contact-state prediction, bounding box detection, and instance segmentation in a comprehensive Hand-Object Segmentation task.
Recent advances have embraced transformer architectures adapted from third-person HOI detection \cite{zou2021end} to capture long-range dependencies and contextual relationships between hands, objects, and scene elements through global-local correlation mechanisms. Simultaneously, synthetic data approaches have emerged to address limited annotation availability, with works like HOI-Synth \cite{LEONARDI2024103984} demonstrating effectiveness across established benchmarks (VISOR, EgoHOS, ENIGMA-51) through systematic domain adaptation strategies. 
Despite these advances, existing approaches face significant computational challenges, particularly for real-time applications. Most current methods employ end-to-end architectures that jointly optimize object detection and interaction understanding, leading to complex models with substantial inference overhead. Transformer-based approaches, while effective at capturing contextual relationships, introduce quadratic computational complexity that limits their deployment in resource-constrained environments. Similarly, multi-modal integration strategies often require sophisticated fusion mechanisms that further increase computational demands. These limitations highlight the need for efficient architectures that can maintain high accuracy while enabling real-time performance.
The challenge of achieving real-time video understanding is not unique to HOI and is a central focus in the related field of Online Action Detection (OAD). OAD seeks to identify an action in a video stream as early as possible from partial observations, making predictions without access to future frames. Recent advancements are largely dominated by Transformer-based architectures like OadTR~\cite{wang2021oadtr} and TeSTra~\cite{zhao2022real} due to their strength in processing long sequences. Alongside these, alternative architectures have demonstrated competitive performance while addressing efficiency concerns. For instance, the RNN-based MiniRoad~\cite{an2023miniroad} offers similar results with a smaller memory footprint, and emerging models have shown success by replacing Transformer components with Mamba-based architectures~\cite{chen2024video}, highlighting the active exploration of different architectural backbones for this real-time task.

In contrast to these approaches, our method addresses both computational and annotation complexity challenges by decomposing egocentric HOI detection into two lightweight components: object detection and contact state estimation. This decomposition enables independent optimization of each module, reducing inference time while requiring minimal annotations compared to comprehensive frameworks demanding segmentation masks, active object vectors, and temporal boundaries \cite{LEONARDI2024103984}. 
Our approach achieves computational efficiency and practical scalability for real-time applications without sacrificing fine-grained interaction understanding.


\begin{figure}[t]
\centering
\includegraphics[width=\textwidth]{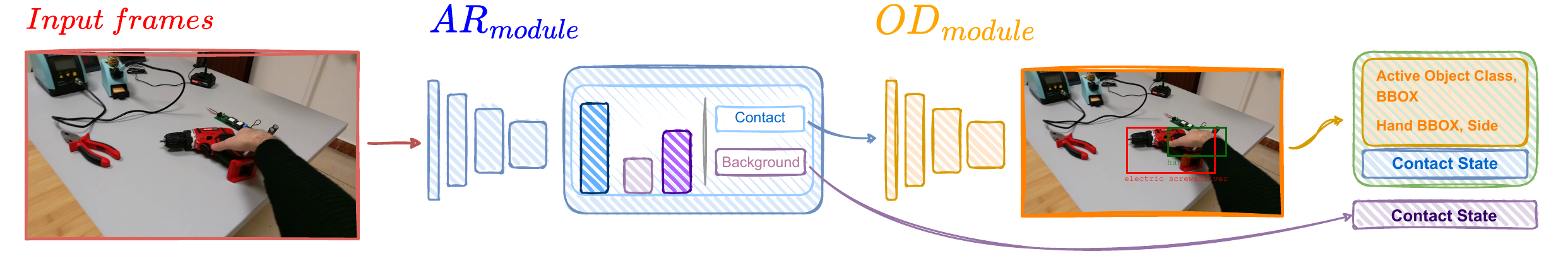}
\caption{Schema of the proposed approach. The AR module predicts the contact state online. Whenever a contact is detected, the object detection module identifies the active object class by evaluating the Intersection over Union (IoU) between objects and hands.} 
\label{fig:approaches}
\end{figure}

\section{Methodologies} \label{approaches}
In this section, we present the pipeline and its components. We detail the integration of the OD and AR modules, describing each component's role and its interaction with the overall framework.

\subsection{Pipeline Overview and Component Details}
We consider two pipelines for detecting hand-object interactions: a baseline and our proposed methodology, illustrated in Figure~\ref{fig:approaches}. Both pipelines employ an OD module to localize hands and objects, but differ fundamentally in how interaction is inferred. The baseline relies on a simple spatial heuristic: after detecting hands and objects using OD, it applies an IoU threshold to infer contact. In contrast, our approach introduces an AR module that first detects potential interaction events. Upon a positive prediction, the AR module activates the OD stage to localize the relevant hand and object. While Figure~\ref{fig:approaches} illustrates only our proposed pipeline, we describe both approaches in detail below, including their respective architectures, training procedures, and integration strategies.
\paragraph{Action Recognition using Mamba.} As AR module, we adopted a Mamba model, which is a modified version of TeSTra \cite{zhao2022real}, with the main goal to predict the action occurring in each frame.
The model predicts a single “contact” action in an online fashion, discriminating it from the background class. The contact class represents those time instants where the hand has just touched something

Our architecture includes a sequence of $3$ Mamba blocks observing the previous 3 seconds of frames. These mamba blocks are preceded by a linear projection of the input features and followed by a classification layer that outputs the classification score, which is followed by a softmax function.
We extracted the frame-wise feature using DINOv2 \cite{oquab2023dinov2} and trained the model using these features as input. 
Since DINOv2 is not suitable for a real-time scenario, we also tested the EfficientNetV2 model~\cite{tan2021efficientnetv2}, distilled from DINOv2.
In particular, we fine-tuned EfficientNetV2 with DINOv2's features as ground truth to force the similarity between the two architectures, with the use of the MSE as loss function.

\paragraph{Object Detection with YOLO.}\label{hoi_explaination} To detect hands and active objects in the scene, we implemented an OD module. We fine-tuned different OD models, such as RT-DETR and YOLOWorld, and selected the best-performing model.
Since we needed a fast and simple solution for online processing, we classified an object as active based on its IoU value with the hand having the highest confidence score in the scene, with the threshold being an hyperparameter. We limited the recognized hands to the two with the highest confidence scores. Additionally, we distinguished left from right hands by considering the position of the bounding box centroid relative to the center of the image's x-axis.

\section{Experimental Settings and Results} \label{results}
We trained our models on the merge of the ENIGMA-51~\cite{ragusa2023enigma} training and validation set. This egocentric dataset provides videos captured in an industrial scenario, including 45,505 RGB frames, 12,597 interaction frames annotated with 14,036 interactions, and 9,342 active objects. We used the dataset to train the RT-DETR and YOLO models, using its fixed taxonomy. 
We used the ENIGMA-51 test set to evaluate the individual modules.
Meanwhile, for the HOI performance evaluation, we used a subset of the ENIGMA-51 test set involved in \cite{LEONARDI2024103984}. This subset contains several ENIGMA-51 clips with a different split. Hence, we filtered out the clips used for training.
All experiments were performed on an NVIDIA Tesla V100S 32GB.

\subsection{Action Recognition Settings and Results}  We formulate the AR task with the constraint to predict the single frame in which the action occurs rather than the segment that defines the start and the end of the action. To accomplish this, we trained our Mamba model for 100 epochs, with an AdamW optimizer, a learning rate of 1e-5 and, a batch size of 32. 
The high framerate (30 fps) of the dataset significantly increase the inherent class imbalance of the annotations. To address this, only during training, we implemented a targeted downsampling strategy, reducing the video framerate from the original 30fps to 4 fps while ensuring all frames with positive annotations were preserved. This approach was complemented by the use of a Focal Loss function, configured to assign a higher weight to the positive class. This combined strategy facilitates a stable training process and, critically, allows the recurrent nature of the Mamba architecture to be leveraged for effective inference on the original high-framerate videos, achieving robust performance.

We evaluated our AR models by measuring the p-AP, as in \cite{shou2018online}. 
Following the metric formulation, given a set of time thresholds, for each ground truth frame, we calculate the nearest prediction greedily favoring high confidence prediction. If the prediction has the same class as the ground truth and the time distance between the two is lower than the time threshold, the prediction is matched to the ground truth. Any unmatched prediction will count as a false positive, while unmatched ground truth instances count as false negatives.
To leverage the powerful feature representations of DINOv2 within a real-time system, we adopted a knowledge distillation strategy. 
The large DINOv2 model serves as the ``teacher''. While EfficientNetV2 acts as the ``student''. The objective is to align the student's feature space with the teacher's one. To achieve this, we used the Mean Squared Error (MSE) as the training loss. 
Due to hardware and time constraints, distilling the model on the entire dataset was computationally prohibitive. To obtain a feasible yet representative training environment, we randomly sampled 10\% of the frames from the complete dataset. This subset was then split into a final training set and a validation set, which were used to effectively train our model and monitor its performance.

The quantitative results for the Mamba-based action recognition module, detailed in Table \ref{tab:testra}, highlight a clear trade-off between accuracy and computational efficiency across different backbones.

The configuration using DINOv2 as a feature extractor establishes a strong performance baseline, achieving an p-AP of $66.28\%$ at 4fps and $40.25\%$ at 30fps. However, with an execution time of $0.06$ seconds per frame, this model's high latency makes it unsuitable for real-time applications at 30fps.

In contrast, employing the standard EfficientNetV2 backbone drastically improves efficiency, reducing the execution time to just 0.007 seconds. This gain in speed comes with a significant performance penalty, as the p-AP drops to $61.32\%$ (from $66.28\%$) at 4fps and to $24.72\%$ (from $40.25\%$) at 30fps.

The distilled EfficientNetV2 model emerges as the optimal configuration. Notably, it not only improves upon the standard EfficientNetV2 but surpasses all other variants at 4fps, achieving a top p-AP of $68.37\%$. While its $30$fps performance does not exceed the DINOv2 baseline, it demonstrates a substantial recovery of accuracy compared to the non-distilled version. Crucially, it achieves this strong performance while maintaining the real-time execution speed of 0.007 seconds, confirming that the distillation process was effective in creating a model that successfully balances high accuracy with low computational overhead.

\begin{table}[t]
\centering
\resizebox{\columnwidth}{!}{%
\begin{tabular}{lcccc}
\toprule
\textbf{Configuration} & \hspace{6pt} \textbf{p-AP@4fps(\%)$\uparrow$}& \hspace{6pt} \textbf{p-AP@30fps(\%)$\uparrow$} & \hspace{6pt} \textbf{Execution Time(s)}$\downarrow$ & \textbf{Size(Mb) $\downarrow$}\\
\midrule
Mamba + DINOv2 & \underline{66.28} & \textbf{40.25} &  0.06 & \underline{448.0}\\
Mamba + EfficientNetV2 & 61.32 & 24.72 & \textbf{0.007} & \textbf{446.0}\\
Mamba + EfficientNetV2 Distilled & \textbf{68.37} & \underline{38.52} & \textbf{0.007} & \textbf{446.0}\\
\bottomrule
\end{tabular}
}
\caption{Comparison of various backbones with Mamba. p-AP results are reported as percentages. Execution Time is reported in seconds (it refers to the time that the complete model needs to process a single frame), model Size in Mb. Best results in bold and second-best results underlined.}
\label{tab:testra}
\end{table}

\subsection{Object Detection Models Comparison}  We tuned and evaluated RT-DETR Large, YOLOv10m, YOLOv8m-world, and YOLOv8x-world. All object detectors were fine-tuned for 200 epochs with 7 epochs of patience, the optimizer was SGD with a learning rate of 0.01 and a momentum of 0.9. The batch size was set to 16, and the input image size to 640x640 without stretching, using padding. The evaluation was done with a confidence threshold of 0.7. 

\begin{table}[t]
\centering
\resizebox{0.7\columnwidth}{!}{%
\begin{tabular}{lcccccc}
\toprule
\textbf{Model} & \hspace{2pt} \textbf{AP(\%) $\uparrow$} & \hspace{6pt} \textbf{Recall(\%) $\uparrow$} & \hspace{6pt} \textbf{HM(\%) $\uparrow$} & \hspace{6pt} \textbf{Execution Time(s) $\downarrow$} & \hspace{6pt} \textbf{Size(Mb) $\downarrow$}\\ 
\midrule
RT-DETR Large &  82.9 & 71.2 & 76.6 & 0.070 & \underline{686.0}\\
YOLOv10m &  83.3 & 70.1 & 76.1 & 0.039 & 754.0\\
YOLOv8m-world &  \textbf{85.1} & \underline{77.7} & \underline{81.2} & \textbf{0.037} & \textbf{634.0} \\
YOLOv8x-world &  \underline{84.8} & \textbf{78.9} & \textbf{81.7} & \underline{0.038} & 862.0\\
\bottomrule
\end{tabular}
}
\caption{Comparison of object detection models. For each row, we report Average Precision (AP), Recall, and Harmonic Mean (HM) in percentages, along with Execution Time in seconds, and model Size (in MB). Best results in \textbf{bold}, second-best in \underline{underlined}.}
\label{tab:obj_quant}
\end{table}

Table~\ref{tab:obj_quant} reports the results of the evaluated models. Among the evaluated models, YOLOv8m-world achieves the highest AP of 85.1\%, making it the most accurate in detecting objects. It is also the most efficient one, with the lowest inference (0.037s) and the smallest model size (634 MB). On the other hand, while YOLOv8x-world has a slightly lower AP (84.8\%), it achieves the highest recall (78.9\%), indicating a stronger ability to detect a wider range of objects. This comes at the cost of a slightly larger model size (862 MB) and a slightly longer inference time. The YOLOv10m model delivers a solid AP of 83.3\%, but it has a slightly lower recall (70.1\%) compared to the YOLOWorld models. While its inference speed (0.039s) is competitive, its model size of 754 MB places it between YOLOv8m-world and YOLOv8x-world in terms of storage efficiency. Meanwhile, RT-DETR Large, despite achieving a respectable AP of 82.9\% and a large size (686.0), has a lower HM value (76.6\%) and a slightly higher latency (0.070s).
\begin{figure}[t]
\includegraphics[width=\textwidth]{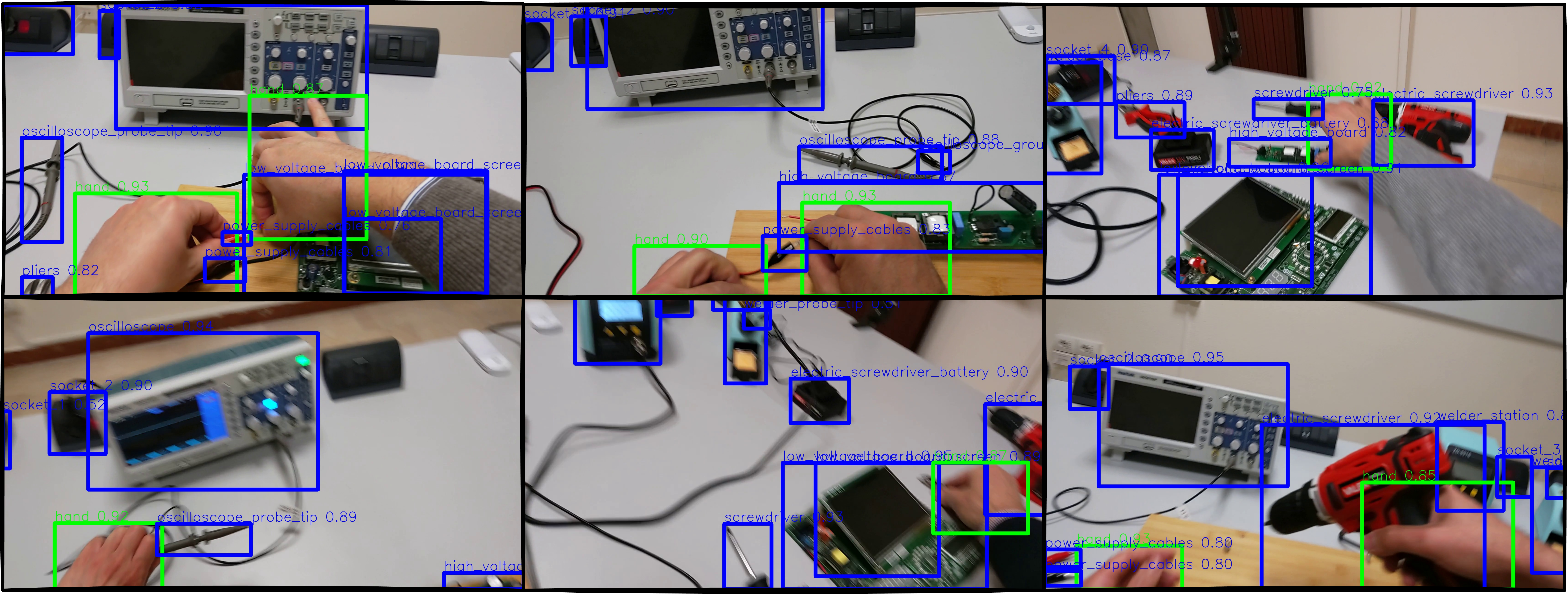}
\caption{Representative samples of YOLOWorldM predictions on some validation images of ENIGMA-51. \textcolor{blue}{Blue} boxes represent objects, \textcolor{green}{green} boxes represent hands.} 
\label{fig:val_samples}
\end{figure}
Figure \ref{fig:val_samples} shows the predictions of some validation images obtained with a confidence score threshold of 0.5.

\subsection{Active Object Retrieval Criteria}
The OD module is integrated into the system with the purpose of identifying the active object once the AR module has detected a contact. As such, the OD module operates conditionally, relying on the output of the AR module. To associate an active object with the hands in the scene, we adopt a straightforward strategy: among all detected objects, the one with the highest IoU with the hand bounding boxes is selected as the active object, provided the IoU exceeds a predefined threshold. Considering no learning is involved, the IoU threshold is the only parameter that requires tuning in this setup.






The performance of the HOI have been evaluate using:
\begin{itemize}
    \item \textbf{AP Hand}: Average Precision of the hand detections.
    \item \textbf{AP Hand + State}: Average Precision of the hand detections considering the correctness of the hand state.
    \item \textbf{AP Hand + Side}: mean Average Precision of the <hand, active object> detected pairs.
    \item \textbf{AP Hand + All}: combinations of AP Hand, AP Hand + State and AP Hand + Side metrics.
\end{itemize}

\begin{table}[t]
\centering
\resizebox{\columnwidth}{!}{%
\begin{tabular}{lcccccc}
\toprule
\textbf{IoU Threshold} & \hspace{8pt}\textbf{AP Hand(\%) $\uparrow$}\hspace{6pt} & \hspace{6pt}\textbf{AP Hand + State(\%) $\uparrow$}\hspace{6pt} & \hspace{6pt}\textbf{AP Hand + Side(\%) $\uparrow$}\hspace{6pt} & \hspace{6pt}\textbf{AP Hand + All(\%) $\uparrow$} \\
\midrule

0.01 &  90.86 & \textbf{47.34} & 88.41 & \textbf{33.90}\\
0.05 &  90.86 & \underline{46.12} & 88.41 & \underline{33.52}\\
0.1 &  90.86 & 35.86 & 88.41 & 28.57\\
0.2 &  90.86 & 23.29 & 88.41 & 20.28\\
0.3 &  90.86 & 19.65 & 88.41 & 14.12\\

\bottomrule
\end{tabular}
}
\caption{Comparison of AP for Hand, Hand + State, Hand + Side, and Hand + All across different IoU thresholds. Best results are shown in \textbf{bold}, and second-best results are \underline{underlined}.}
\label{tab:quantres}
\end{table}

The results reported in Table \ref{tab:quantres} were computed only on frames where contact has been annotated. Under this assumption, we activate the OD module through  an oracle that does not penalize false positives detected by the module. As shown, the best results are obtained with an IoU threshold of 0.01. This low threshold requirement highlights the need for a more advanced method to detect interactions.

\begin{figure}[t]
\includegraphics[width=\textwidth]{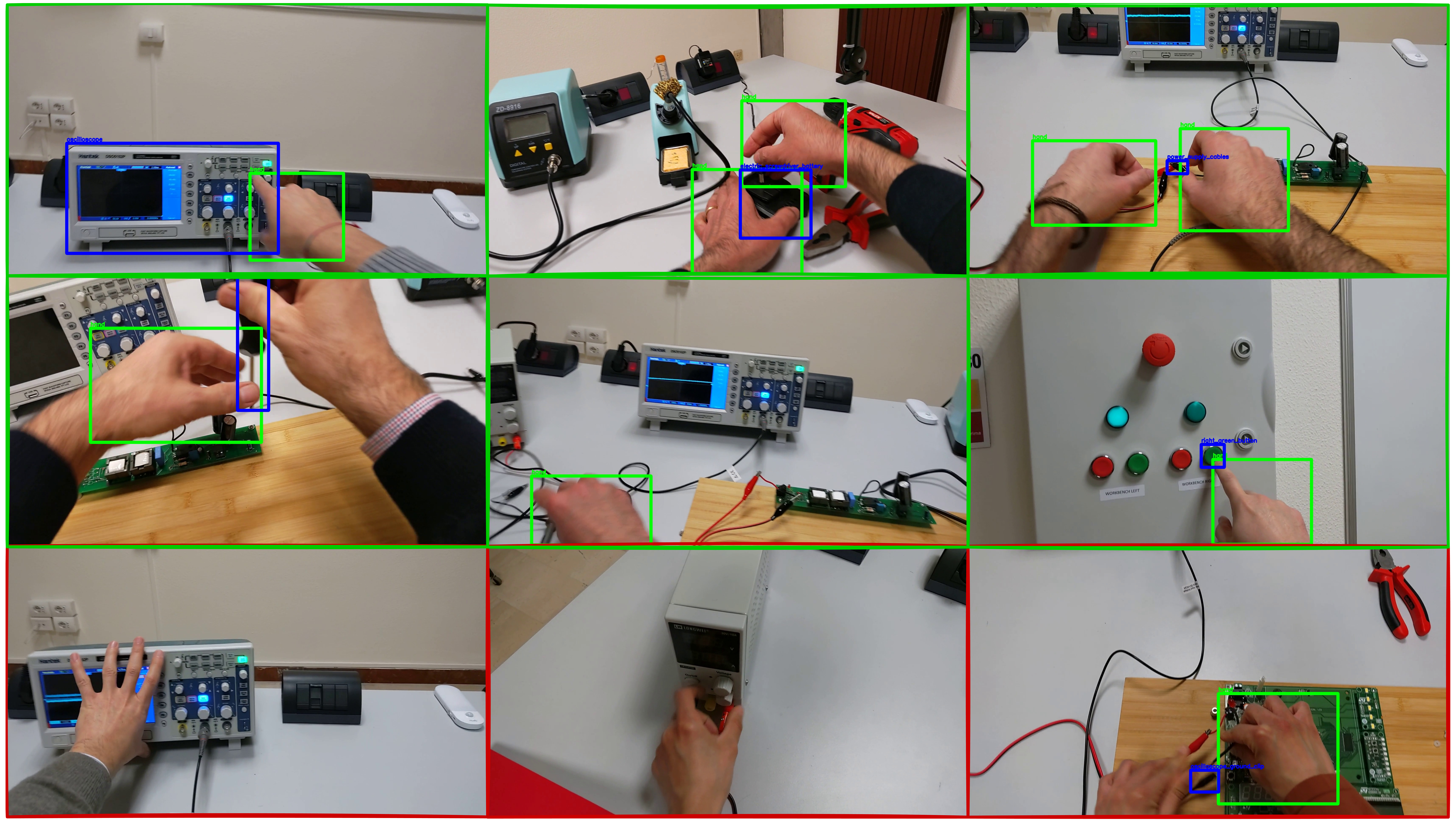}
\caption{Representative samples of active object retrieval performed by YOLOWorld. \textcolor{green}{Green} frames highlight success cases, \textcolor{red}{red} frames indicate failure cases.} 
\label{fig:qual}
\end{figure}

Considering that Mamba has no integrated mechanism to retrieve the class of the object, we decided to use YOLOWorld as a model to retrieve the active object in the scene and use Mamba to detect the action. 
Figure \ref{fig:qual} illustrates some ENIGMA-51 test samples. As can be seen, the algorithm returns one object per frame, choosing between objects and hands based on the highest IoU. The frames inside the green boxes indicate successes, where YOLO correctly detects the hand and the objects, and the active object prediction is correct. 
Red boxes indicate failures and can be due to various reasons, such as objects or hands not being recognized by YOLO, IoU threshold not being met, or multiple objects close together, leading to incorrect object selection.


\subsection{Overall Pipeline Evaluation}

\begin{table}[t]
\centering
\resizebox{\columnwidth}{!}{%
\begin{tabular}{lcccccc}
\toprule
\textbf{Method} & \hspace{6pt} \textbf{AP Hand(\%) $\uparrow$}\hspace{6pt} & \hspace{6pt}\textbf{AP Hand + State(\%) $\uparrow$}\hspace{6pt} & \hspace{6pt}\textbf{AP Hand + Side(\%) $\uparrow$}\hspace{6pt} & \hspace{6pt}\textbf{AP Hand + All(\%) $\uparrow$}\hspace{6pt} & \textbf{Execution Time(s)}$\downarrow$ \hspace{6pt} & \textbf{Size(Mb) $\downarrow$}\\
\midrule

ORACLE+YOLO &  \textbf{90.86} & \textbf{47.34} & \textbf{88.41} & \textbf{33.90} & \textbf{0.057} & \textbf{634.0}\\
MAMBA@30+YOLO &  36.34 & 23.22 & 35.36 & 14.98 & \underline{0.064} & \underline{1080} \\
MAMBA@60+YOLO &  \underline{45.43} & \underline{28.85} & \underline{44.37} & \underline{19.70} & \underline{0.064} & \underline{1080}\\

\bottomrule
\end{tabular}
}
\caption{Comparison of AP for Hand, Hand+State, Hand+Side, and Hand+All using ORACLE+YOLO and MAMBA-based pipelines. ORACLE+YOLO serves as an upper bound using ground-truth labels, while MAMBA@30/60+YOLO leverages predictions from the past 30 and 60 frames, respectively. Best results are in \textbf{bold}, second-best in \underline{underlined}.}
\label{tab:overall}
\end{table}

We evaluate two strategies for triggering object detection (OD) in the context of hand-object interaction (HOI): a heuristic oracle-based method and a learned approach using the AR module.
In the oracle-based setup, we assume perfect knowledge of when a contact occurs and trigger the OD module (YOLOv8m-world) only on frames known to contain contact. Within these frames, the active object is identified using the aforementioned mechanism involving bounding boxes overlap. Despite being driven by a simple heuristic, this method establishes an upper bound on performance, as it leverages ground-truth contact timing.
In contrast, the Mamba-based AR module detects contact events over time and autonomously triggers the OD module. For online processing, we apply a temporal windowing strategy that looks only at past frames: if the AR module detects contact within the preceding 30 or 60 frames from the current frame, the OD is triggered for that frame. This past-only approach ensures real-time compatibility while reducing sensitivity to occasional missed detections and better capturing the temporal nature of interactions.

Table~\ref{tab:overall} compares the results. The first row (ORACLE+YOLO) represents the upper bound. The second and last rows show the results when using our AR model with different temporal windows. Notably, the oracle setting does not penalize false positives from YOLO, as OD is invoked only in frames known to contain contact. This leads to higher reported values compared to the AR-triggered setup, where the OD is invoked based on predicted contact, potentially introducing more uncertainty. A more balanced comparison would require accounting for false positives in the oracle-driven YOLO as well, which would likely narrow the performance gap.

\section{Implementation with Meta Quest 3}
 
\begin{figure}[t]
\includegraphics[width=\textwidth]{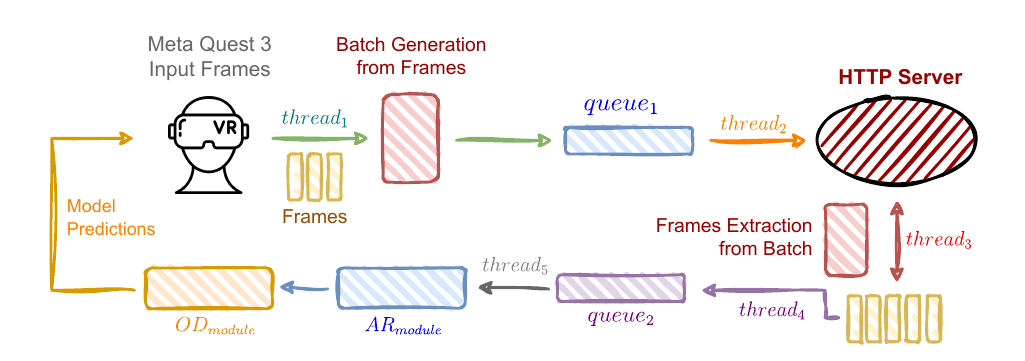}
\caption{Representation of the full pipeline, from input frames recording to model predictions back to the Meta Quest 3.} 
\label{fig:pipe-meta}
\end{figure}

In this section, we present a full pipeline that leverages a wearable device to capture frames and acts as a practical example.
To operate in a real-time fashion, we utilize the Meta Quest 3 headset as the source for video streaming and employ multiple threads to handle processes concurrently. All frames are captured via a Unity-based interface, enabling the user to start and stop the recording using the Meta Quest controller.
When the recording begins, the Meta Quest stores up to 60 frames. $thread_1$ is responsible for creating a batch of frames and inserting them into $queue_1$. Subsequently, $thread_2$ retrieves the batch and sends it to an HTTP server.
The HTTP server receives the batch, and $thread_3$ unpacks it into 60 individual frames. From this, $thread_4$ extracts the frames, duplicates each one, and sends them into $queue_2$, which facilitates the flow of frames to the models. $thread_5$ serves as the model worker for $queue_2$; it extracts frames in groups of 30, sends them to the models for concurrent predictions, and finally returns the results to the Meta Quest 3 for user feedback. The explained pipeline is illustrated in Figure \ref{fig:pipe-meta}.
Regarding the Unity application used to send the frames and visualize user feedback, our work is based on the Android MediaProjector API. This allows us to record the left eye screen of the Meta Quest 3 and access the raw frames.
We adopted this method because we believe that an ``edge-based'' solution like this one is superior to casting.
In fact, the alternative is to use the casting option of Meta and than record the browser: this solution, even if it is faster in latency when operating under optimal solution, is not viable if we consider that it is mandatory that both the visor and the computer needs to be in the same network. With our solution, by adopting an HTTP request to an arbitrary endpoint, we can send those frames independently of the network used for both of the elements included.

\section{Conclusion}
In this work, we tackled the pressing challenge of detecting hand-object interactions from an egocentric perspective in real-time, a critical capability for industrial applications. We introduced an efficient, two-stage approach that decomposes this complex problem into online contact detection and active object identification. Our proposed pipeline first leverages an AR module to determine when contact occurs, which in turn triggers an OD module to identify the specific hand and object involved in the interaction.
Our experimental results validate the effectiveness of this methodology. For the AR task, we demonstrated that a Mamba-based model, when paired with a distilled EfficientNetV2 feature extractor, successfully balances high accuracy with the low latency required for real-time applications, processing frames in just 0.007 seconds. For the OD component, YOLOWorld emerged as the superior model, achieving the highest accuracy ($85.13\%$ AP) and fastest inference speed ($0.037s$) among the tested alternatives.
Our analysis revealed that a very low IoU threshold was required for optimal performance, even with oracle contact data, highlighting the limitations of inferring interactions from spatial overlap alone.
The subsequent performance decrease when combining the Mamba-based Action Recognition module and YOLOWorld is justified, as the system transitions from an idealized oracle to a practical, learned model. The oracle setup represents a theoretical upper bound on performance because it uses ground-truth data to trigger object detection only on frames with guaranteed contact, a method that is not penalized for false positives.
Finally, we demonstrated the practical applicability of our work by developing and implementing the entire pipeline on a Meta Quest 3 headset.

Code and implementation details are available at: \url{https://github.com/JustCati/iCODE-Server}.


\section*{Acknowledgements}
This research has been funded by the European Union - Next Generation EU, Mission 4 Component 1 CUP E53D23008280006 - Project PRIN 2022 EXTRA-EYE, and FAIR – PNRR MUR Cod. PE0000013 - CUP: E63C22001940006. A. Finocchiaro has been supported by Research Program PIAno di inCEntivi per la Ricerca di Ateneo 2024/2026, project "Multi-Agent Simulator for Real-Time Decision-Making Strategies in Uncertain Egocentric Scenarios" - University of Catania.


\bibliographystyle{splncs04}
\bibliography{MixedICIAP}

\end{document}